%% file: aaai25.tex
\documentclass[letterpaper]{article} 
\usepackage{aaai25}  
\usepackage{times}  
\usepackage{helvet}  
\usepackage{courier}  
\usepackage[hyphens]{url}  
\usepackage{graphicx} 
\usepackage{natbib}  
\usepackage{caption} 
\usepackage{algorithm}
\usepackage{algorithmic}

\usepackage{newfloat}
\usepackage{listings}
\DeclareCaptionStyle{ruled}{labelfont=normalfont,labelsep=colon,strut=off} 
\floatstyle{ruled}
\newfloat{listing}{tb}{lst}{}
\floatname{listing}{Listing}
\title{Attention-driven GUI Grounding: Leveraging Pretrained Multimodal Large Language Models without Fine-Tuning}
\author{
    Hai-Ming Xu\textsuperscript{\rm 1}\equalcontrib,
    Qi Chen\textsuperscript{\rm 1}\equalcontrib,
    Lei Wang\textsuperscript{\rm 2},
    Lingqiao Liu\textsuperscript{\rm 1}\thanks{Corresponding Author.}
}
\affiliations{
    \textsuperscript{\rm 1}Australian Institute for Machine Learning, The University of Adelaide\\

    \textsuperscript{\rm 2}University of Wollongong\\
    \{hai-ming.xu, qi.chen04, lingqiao.liu\}@adelaide.edu.au, lei\_wang@uow.edu.au
}

\usepackage{bibentry}

\usepackage[table]{xcolor}
\usepackage{url}
\usepackage{graphicx}
\usepackage{amsmath}

\usepackage{makecell}
\usepackage{arydshln}

\usepackage{cleveref}
\usepackage{multirow}
\usepackage{amssymb}
\usepackage{bbding}
\usepackage{pifont}
\definecolor{darkred}{rgb}{0.8, 0, 0}  
\definecolor{darkgreen}{rgb}{0, 0.5, 0} 
\usepackage{colortbl}
\usepackage{float}
\usepackage{array}
\newcolumntype{C}[1]{>{\centering\arraybackslash}m{#1}}

\begin{document}

\maketitle

\begin{abstract}
Recent advancements in Multimodal Large Language Models (MLLMs) have generated significant interest in their ability to autonomously interact with and interpret Graphical User Interfaces (GUIs). A major challenge in these systems is grounding—accurately identifying critical GUI components such as text or icons based on a GUI image and a corresponding text query. Traditionally, this task has relied on fine-tuning MLLMs with specialized training data to predict component locations directly. However, in this paper, we propose a novel Tuning-free Attention-driven Grounding (TAG) method that leverages the inherent attention patterns in pretrained MLLMs to accomplish this task without the need for additional fine-tuning. Our method involves identifying and aggregating attention maps from specific tokens within a carefully constructed query prompt. Applied to MiniCPM-Llama3-V 2.5, a state-of-the-art MLLM, our tuning-free approach achieves performance comparable to tuning-based methods, with notable success in text localization. Additionally, we demonstrate that our attention map-based grounding technique significantly outperforms direct localization predictions from MiniCPM-Llama3-V 2.5, highlighting the potential of using attention maps from pretrained MLLMs and paving the way for future innovations in this domain. Code is available at \url{https://github.com/HeimingX/TAG.git}.
\end{abstract}

%

\input{contents/1-introduction}
\input{contents/2-related_works}

\input{contents/3-method}
\input{contents/4-experiments}
\input{contents/5-conclusion}

\bigskip
\paragraph{Acknowledgements}
This work was supported by the Centre for Augmented Reasoning, an initiative by the Department of Education, Australian Government.

\input{contents/7-supp}

\bibliography{aaai25}

\end{document}

%% file: contents/1-introduction.tex
\section{Introduction}\label{sec:intro}
The integration of artificial intelligence with Graphical User Interfaces (GUIs) holds tremendous potential to transform how humans interact with software systems. Leading this innovation are Multimodal Large Language Models (MLLMs)~\cite{gpt4v,reid2024gemini,claude2024}, which have shown exceptional capabilities in interpreting GUIs across various applications. A crucial task in AI for GUIs is GUI grounding—accurately identifying and localizing key components such as text and icons—since this is fundamental to enabling the automated operation of GUIs. While MLLMs excel at understanding GUI images, the precise grounding of GUI elements remains challenging.

Current state-of-the-art solutions often improve the GUI grounding capabilities of MLLMs through fine-tuning on specialized datasets, as demonstrated in works like~\cite{hong2023cogagent,cheng2024seeclick}. In these methods, the MLLM directly predicts the location of GUI elements. In contrast, our approach takes a different path by leveraging the inherent attention patterns of a pretrained MLLM, utilizing its existing spatial awareness and attention mechanisms to achieve accurate GUI grounding without the need for additional fine-tuning.

We propose a novel Tuning-free Attention-driven Grounding (TAG) approach that carefully selects and aggregates attention patterns from MiniCPM-Llama3-V 2.5, a state-of-the-art MLLM, to perform GUI element grounding. Our method begins by identifying specific tokens from either the user input query or the model-generated response and then propagates the corresponding attention values back to the image plane. To further enhance performance, we implement a selective mechanism to filter out irrelevant attention heads, ensuring that only the most relevant attention is utilized for accurate grounding.

We compare our approach with existing tuning-based methods, and our results demonstrate that utilizing attention patterns from a pretrained model can achieve accurate GUI element grounding. Additionally, our approach significantly improves text localization. These findings highlight the untapped potential of leveraging inherent model capabilities and open the door to more robust, scalable, and efficient applications of MLLMs in GUI automation.

\input{figures/fig_reason_well_loc_fail}

%% file: figures/fig_reason_well_loc_fail.tex
\begin{figure*}[t!]
\centering
\includegraphics[width=0.85\textwidth]{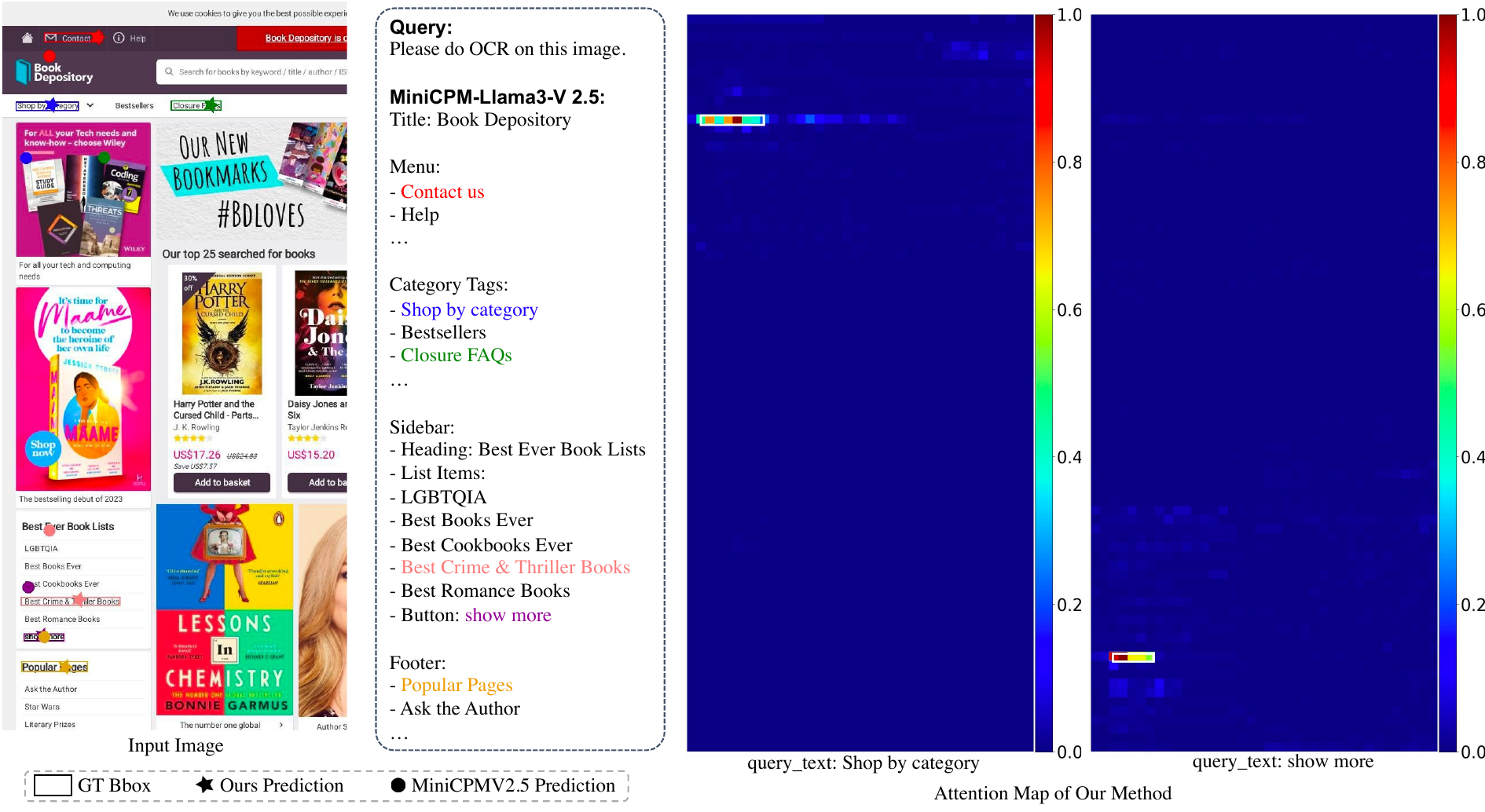}
\captionsetup{skip=2pt}
\caption{
Illustration of MiniCPMV2.5's strong GUI image understanding but poor element localization. Our attention-driven GUI grounding leverages its inherent attention to enhance localization accuracy without fine-tuning, as shown on the right.
}
\label{fig:reason_well_loc_fail}
\vspace{-.35cm}
\end{figure*}

%% file: contents/2-related_works.tex
\section{Related Works}

\paragraph{Multimodal Large Language Models for GUI Agents}
The use of MLLMs~\cite{liu2023llava,qwenvl,yao2024minicpmv,liu2024visual,zhu2023minigpt,lu2024deepseekvl,wang2023cogvlm} as GUI agents marks significant progress in AI's ability to interact with GUI. These models understand user queries and images, enabling them to perform tasks across various platforms, from desktops to mobiles. Recent works in this domain have explored various applications, from automating routine tasks on desktop interfaces~\cite{hong2023cogagent,wu2024copilot,kil2024dual,he2024webvoyager,kapoor2024omniact,xie2024osworld} to providing interactive assistance on mobile platforms~\cite{ma2024coco,nong2024mobileflow,wang2024mobile,wang2024mobileagentbench,you2024ferret}. These applications highlight the potential of MLLMs to act as autonomous agents that can understand and execute user commands across different platforms. However, the challenge often lies in effectively training these models to handle the intricacies and variability of GUIs without extensive domain-specific tuning.

\paragraph{Grounding in GUI Agents}
Grounding in GUI agents~\cite{cheng2024seeclick,li2020interactive,liu2024visualwebbench,gao2024assistgui,wang-etal-2024-devils,li2023spotlight,conv_wang2023,li-etal-2020-mapping} involves the model's ability to locate and identify interface elements accurately, which is essential for effective interaction. Traditional methods~\cite{cheng2024seeclick,chen2024guicourse,hong2023cogagent,fan2024read} typically require fine-tuning on detailed, annotated datasets. Recent research has explored both supervised and unsupervised techniques to enhance grounding accuracy, such as the SeeClick model~\cite{cheng2024seeclick} fine-tunes on GUI-specific datasets. However, these methods can suffer from scalability issues and overfitting. Our work contributes to this field by proposing a tuning-free approach that leverages pre-trained MLLMs' inherent attention mechanisms to associate text queries with visual elements, offering a scalable and adaptable grounding solution.

%% file: contents/3-method.tex
\section{Our Method}
\input{figures/fig_method}

\subsection{Preliminary}\label{sect: preliminary}

\paragraph{GUI Grounding}
GUI grounding is a crucial task for agents that interpret and interact with graphical user interfaces (GUIs). It demands that systems comprehend users' text queries, such as ``I want to book a dental appointment on Tuesday'', analyze GUI screenshots, and accurately pinpoint the relevant components. While recent advancements in MLLMs have shown potential in understanding both textual queries and visual GUI layouts, they often encounter challenges in precisely localizing elements without the aid of additional tools like~OCR \cite{wang2024mobile} or Set-of-Mark~\cite{yang2023set,wang2024mobile} techniques. Thus, SOTA methods typically rely on fine-tuning MLLMs with specialized training data to directly achieve accurate element localization~\cite{hong2023cogagent,cheng2024seeclick}.

\paragraph{MiniCPMV2.5 and Its Attention Map}
MiniCPMV2.5 (i.e., MiniCPM-Llama3-V 2.5) is a SOTA MLLM that integrates a vision encoder, a token compression module, and the Llama3 language model. It supports high-resolution images up to 1344×1344 pixels with any aspect ratio, making it well-suited for precise GUI grounding tasks. To manage the large number of visual tokens generated from high-resolution inputs, the model uses cross-attention to compress thousands of vision patch embeddings into a fixed-size ($Q$) set of visual query tokens. These visual query tokens are then processed alongside text tokens by Llama3, which fuses the two modalities through multi-layer transformers utilizing multi-head self-attention. For more details, refer to the report by~\cite{yao2024minicpmv}.

Empirically, as shown in Fig.~\ref{fig:reason_well_loc_fail}, when presented with a GUI image, MiniCPMV2.5 demonstrates a strong ability to comprehend the UI layout and accurately recognize optical characters within the image. Additionally, due to its training on object-detection-related tasks, MiniCPMV2.5 is capable of localizing objects by predicting the bounding box of the object of interest.

Our method aims to further enhance localization performance by leveraging the attention maps within MiniCPMV2.5. Specifically, MiniCPMV2.5 consists of two major components: the token compression module and the Llama3 LLM, from which attention weights can be extracted. For the token compression module, attention values can be obtained from the cross-attention layer. By averaging these attention weights across all heads, we obtain an attention map $A_{\text{cross}} \in [0,1]^{Q \times H \cdot W}$, where $Q$ is the number of visual query tokens, and $H$ and $W$ are the height and width of the patchified image, respectively. The self-attention weights in Llama3 LLM can be represented as $A_{llm} \in [0,1]^{N \times M \times M}$, where $N$ is the total number of multi-head self-attention (MHA) layers multiplied by the number of attention heads per MHA, and $M$ is the number of tokens input to the LLM, including both visual tokens and text tokens.

\subsection{Overview of Our Method}
Our method focuses on selecting and aggregating attention weights from MiniCPMV2.5 to achieve accurate localization of GUI elements. The key insight of our approach is that a well-crafted selection and aggregation strategy is essential for success. Specifically, our method comprises the following three components:
\begin{enumerate}
    \item \textbf{Adaptive Text Token Selection:}$A_{llm}$ contains self-attention values for all token pairs, but not all of them contribute to effective grounding. This component focuses on identifying the attention between the most relevant tokens to ensure accurate localization.
    \item \textbf{Attention-driven GUI Grounding:} This component aggregates both $A_{llm}$ and $A_{cross}$ to identify the element localization.
    \item \textbf{Self-Attention Head Selection:} This component improves grounding accuracy by selecting high-quality attention heads among 1024 attention heads in Llama3.
\end{enumerate}

\subsection{Adaptive Text Token Selection}\label{subsec:text_token_selection}
GUI grounding tasks aim to locate elements relevant to a user's query. However, user queries often contain numerous tokens, not all of which pertain to the target GUI element. Some queries explicitly identify the element of interest, like ``go to the next page'' implying a click on the 'next page' button, while others only imply it, such as ``take a photo as input'' indirectly referring to the 'Camera' button in the GUI. Figure~\ref{fig:mind2web_agent} illustrates how complex, multi-step queries can struggle to align with dynamic UI changes, leading to inaccurate grounding. Therefore, it's essential to develop a mechanism that selects key tokens and leverages the relevant self-attention weights for accurate GUI grounding.

Leveraging MiniCPMV2.5's remarkable ability to comprehend GUI images, in this paper we propose a simple yet effective strategy: \textit{constructing the query prompt to prompt the model to first explicitly generate a description of the content or elements relevant to the query. We then use the attention between these descriptive tokens $\{\mathcal{T}_j\}_{j=1}^\text{T}$ and visual tokens to achieve localization.} This approach significantly improves GUI grounding performance by bridging the gap between user queries and UI elements.

\input{figures/fig_topk}

\subsection{Attention-driven GUI Grounding}\label{subsec:tag_pipeline}
As discussed in Section \ref{sect: preliminary}, the image tokens are not directly fed into the LLM. Instead, they are first compressed into visual query tokens before being passed to the LLM. This means that the selected text tokens, which correspond to the content or description of the target GUI element, may not directly attend to the image region. To address this issue, we propose a method to propagate attention from the selected text tokens to the image grid. Specifically as illustrated in Fig.~\ref{fig:method}, we leverage the selected text tokens $\{\mathcal{T}_j\}_{j=1}^\text{T}$ from Sec.\ref{subsec:text_token_selection} to generate head-wise attention maps $A'_{llm}\in[0, 1]^{\text{N}\times\text{T}\times\text{Q}}$, which represent the attention between these text tokens and visual query tokens across all layers' multi-head self-attentions in Llama3. Here, $\text{N}$, $\text{T}$, and $\text{Q}$ denote the number of heads in Llama3, the number of selected text tokens, and the number of visual query tokens, respectively.
To obtain an overall relationship between each selected text token and the visual query tokens, we aggregate the attention from different heads by weighted summation: 
\begin{equation}
    \bar{A}_{llm}(\mathcal{T}_j) = \frac{1}{N}\sum_{k=1}^{N} \alpha_{k,j} A'_{llm}[k,j,:] \in [0, 1]^{\text{Q}},
\end{equation}where $k$ is the head index and $\alpha_{k,j}$ is the aggregation weight for $k$-th head and $j$-th selected text token, respectively. The strategy of how to set $\alpha_{k,j}$ will be discussed in Section \ref{subsec:attn_filtering}.
After obtaining $\bar{A}_{llm}(\mathcal{T}_j)$, which represents the attention of each selected text token to each visual query token, we propagate the attention from each visual query token to the corresponding image patch token using $A_{cross}\in[0, 1]^{\text{Q}\times(\text{H}\cdot\text{W})}$. This is accomplished through a simple matrix multiplication:
\begin{equation}
    R_j = \bar{A}_{llm}(\mathcal{T}_j) \times A_{cross} \in [0, 1]^{\text{H}\cdot\text{W}}.
\end{equation} Intuitively, this operation distributes the attention received by each visual query token to the corresponding image patch tokens, proportional to the attention values between the visual query token and each image patch token.
Finally, to obtain an overall relationship between the query text and image patches, we average across different selected text tokens:
\begin{equation}
    \bar{R} = \frac{1}{\text{T}}\sum_{j\in{\{1, 2, \cdots, \text{T}\}}} R_j \in [0, 1]^{\text{H}\cdot\text{W}}.
\end{equation}$\bar{R}$ represents the relevance of image patches to the query. To achieve pixel-level localization, we first map the relevance score from the patch to the pixel level by assigning the same value to all pixels within a patch (e.g., an $14 \times 14$ pixel grid). Next, we apply a threshold $\delta$ to binarize the image and identify connected regions. The region with the highest average relevance score is selected, and its center is used as the predicted location.

\subsection{Self-Attention Head Selection}\label{subsec:attn_filtering}
Empirically, we find that not all self-attention heads in the LLM part  of MiniCPMV2.5 are equally useful for aligning the text tokens to the image patches. As the investigation presented in Fig.~\ref{fig:topk}, to ground the text ``\texttt{Search artists, albums and more}'' in the input field, the method with naive averaging attention maps of all heads falsely ground to the search icon. To find the reason, we further use each head's self-attention to map every text token to the image space separately. As figures shown on the right side of Fig.~\ref{fig:topk}, there are always some attention heads (which are colored in blue) that map the text token outside the ground truth bounding box for every text token, which means not all attention heads corresponding to each text token is equally effective in accurately mapping the token to its expected region. To determine the quality of attention heads, we find that magnitude of the average attention between a selected text token $\mathcal{T}_j$ and visual query tokens can be a good indicator, namely,
\begin{equation}
    \tilde{A}_{\mathcal{T}_j}^{k} = \sum_{q\in\{1,2,\cdots,\text{Q}\}} A'_{llm}[k, j, q].
\end{equation}
This is demonstrated by the observation that when $\tilde{A}_{\mathcal{T}_j}^{k}$ is larger, the head's attention is more likely to map the text token to the intended region. As illustrated in Figure~\ref{fig:topk}, heads with high $\tilde{A}_{\mathcal{T}_j}$ tend to make predictions within the ground-truth bounding box (which are colored in red). This insight leads us to retain only the attentions of heads corresponding to the top-$K$ values of $\tilde{A}_{\mathcal{T}_j}$. Additionally, we observe that the head rankings based on $\tilde{A}_{\mathcal{T}_j}$ vary across different text tokens. Therefore, we select the top heads for each token individually. This strategy effectively sets $\alpha_{k,j}$ to '1' for the selected heads while assigning '0' to the others.

%% file: figures/fig_method.tex
\begin{figure*}[t!]
\centering
\includegraphics[width=0.85\textwidth]{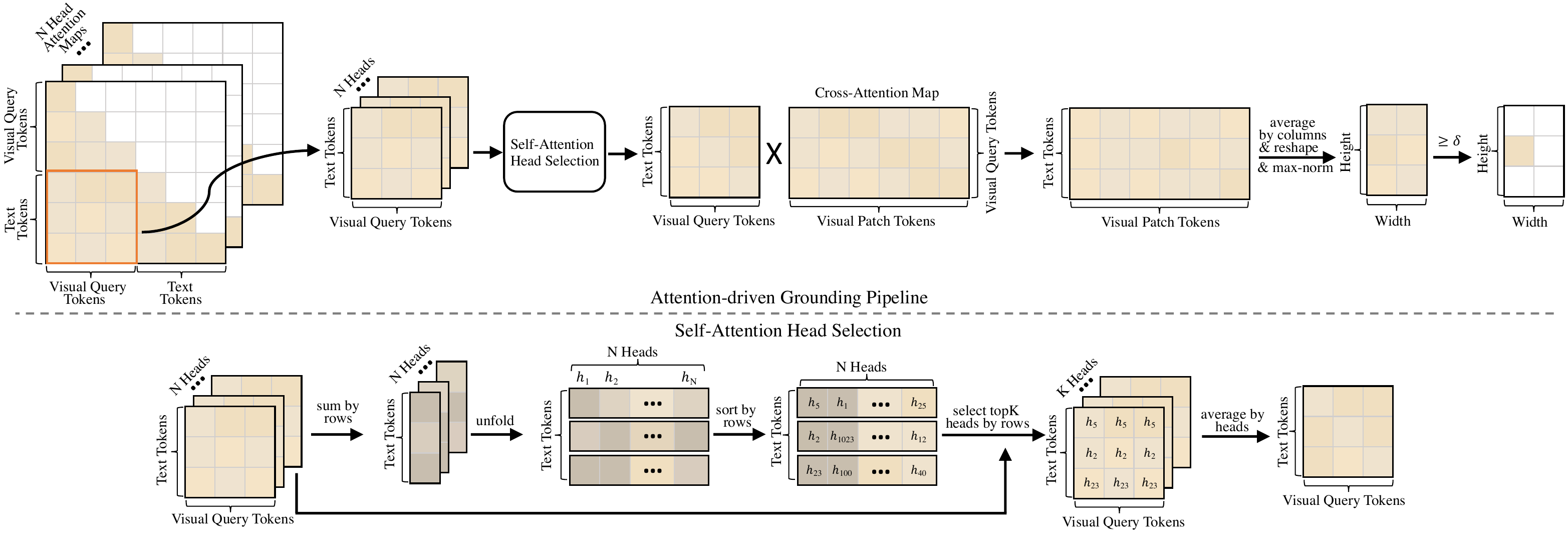}
\captionsetup{skip=2pt}
\caption{Overall pipeline of our TAG approach in Sec.~\ref{subsec:tag_pipeline} (top) and the self-attention selection module in Sec.~\ref{subsec:attn_filtering} (bottom).}
\label{fig:method}
\vspace{-.35cm}
\end{figure*}

%% file: figures/fig_topk.tex
\begin{figure*}[t!]
\centering
\includegraphics[width=0.8\textwidth]{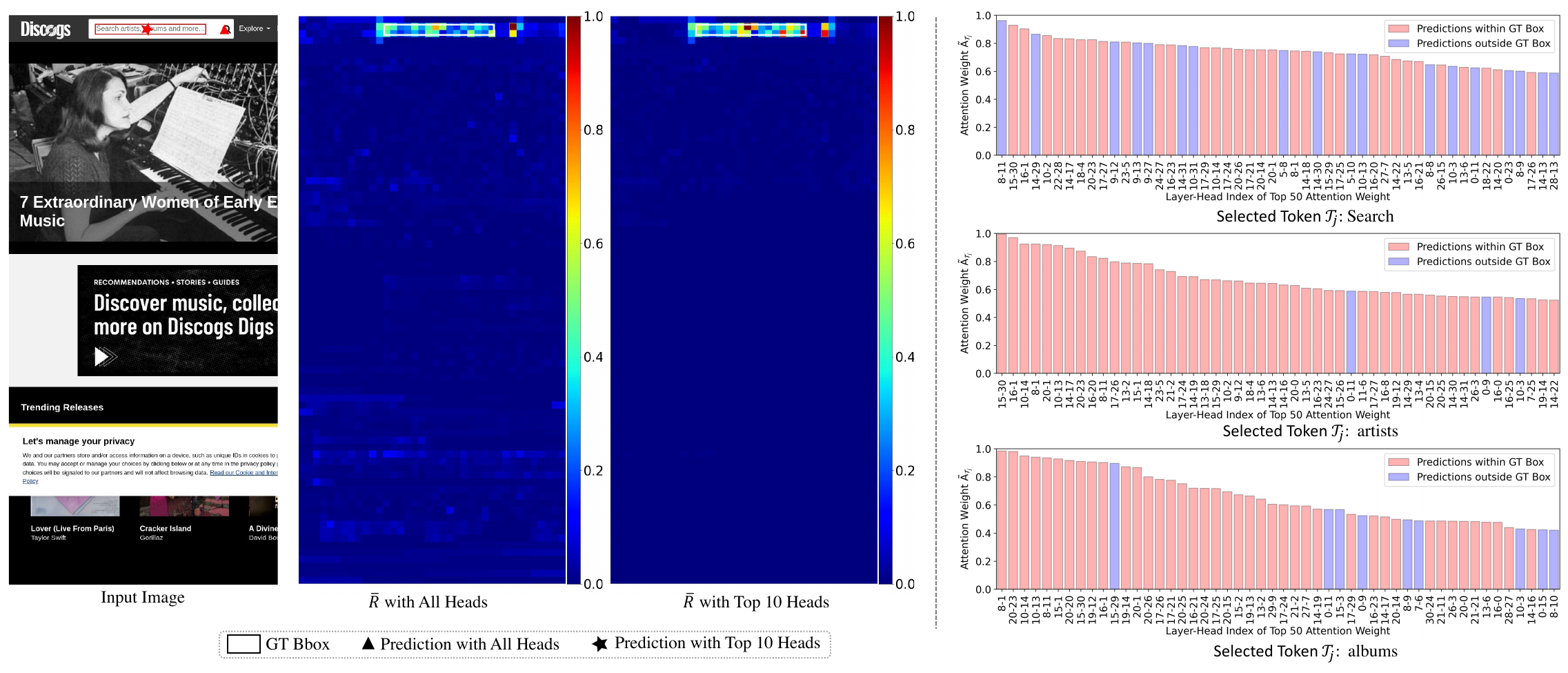}
\captionsetup{skip=2pt}
\caption{
Demonstrating how choosing top self-attention heads improves text-to-image token mapping (see Sec.~\ref{subsec:attn_filtering} for details).
}
\label{fig:topk}
\vspace{-.35cm}
\end{figure*}

%% file: contents/4-experiments.tex
\section{Experiments}
In this section, we compare our method to the SOTA ones on three benchmarks, each designed to test our method from different perspectives. Besides, we conduct several ablation studies to further analyze the effectiveness of our method.
We use the greedy generation strategy in our method for a reproducible result and  
all experiments can be conducted on one NVIDIA RTX 4090 GPU.

\input{tables/tab_m2w_loc_v1}
\input{tables/tab_screenspot}
\input{figures/fig_screenspot}

\subsection{Task1: Optical Character Grounding}
Our method primarily achieves grounding by mapping text tokens to the image space. To directly validate our approach, we developed an optical character grounding benchmark using the Mind2Web~\cite{deng2024mind2web} dataset. While Mind2Web was originally designed for text-based (HTML) GUI agent evaluation in website environments, it also includes corresponding screenshots, which we leveraged to create our novel dataset, \textit{OCG}.

\paragraph{OCG Dataset}
First, we collect homepage screenshots from 104 websites in the Mind2Web test set. We then use the Azure Vision API tool\footnote{https://azure.microsoft.com/products/ai-services/ai-vision} to obtain OCR information for each screenshot. This API can identify all text in the screenshot, including non-element text within images, allowing us to evaluate the MLLM's ability to locate general text.
Besides, to assess model performance across various image aspect ratios, we crop sub-images from the homepage screenshots corresponding to different aspect ratios. We retain only the OCR bounding boxes that fell entirely within these sub-images for evaluation. Based on common screen resolutions\footnote{https://gs.statcounter.com/screen-resolution-stats}, we construct 10 different aspect ratios (width: height): 1:4, 9:21, 9:19, 1:2, 9:16, 4:3, 16:9, 2:1, 21:9, and 4:1. This diverse set of aspect ratios allows us to comprehensively assess our model's robustness to varying image dimensions, which is crucial for real-world applications where screen sizes and orientations can vary significantly.

\paragraph{Baseline Methods} 

We benchmark our approach against three notable models: MiniCPMV2.5~\cite{yao2024minicpmv}, a recently open-sourced SOTA MLLM that serves as the foundation for our method; SeeClick~\cite{cheng2024seeclick}, the current SOTA GUI grounding method; and Qwen-VL-Chat~\cite{qwenvl}, the foundation model for SeeClick. For each model, we use specific prompts tailored to their respective functionalities. For Qwen-VL-Chat, we use ``\texttt{Generate the bounding box of \{query\_text\}}''. SeeClick's prompt is ``\texttt{In this UI screenshot, what is the position of the element "\{query\_text\}" (with point)?}''. MiniCPMV2.5 utilizes the prompt ``\texttt{What is the bounding box of "\{query\_text\}" in the image? The bounding box output format is: <box>xmin ymin xmax ymax</box>. Please directly output the bounding box.}''\footnote{MiniCPMV2.5 sometimes fails to generate the box without the last prompt sentence.}. For our method, we employ ``\texttt{What is the bounding box of "\{query\_text\}"}''. Since the query text is extracted by OCR which is well aligned with the corresponding text in the image, we thus directly use the query text for grounding to verify our method.

\input{figures/fig_mind2web_agent}
\input{tables/tab_m2w_agent_v2}

\paragraph{Results} 
As shown in Tab.~\ref{tab:mind2web_ocg}, the foundation MLLM Qwen-VL-Chat, while capable of detecting general objects, struggles to localize query text in the OCG task. In contrast, the more recent MiniCPMV2.5 demonstrates improved text grounding ability. However, MiniCPMV2.5's performance varies considerably across different aspect ratios, achieving 80.2\% accuracy on the 4:3 aspect ratio but only 13.6\% on 9:21. we speculate that although the model can support inputs of any aspect ratios, its pre-training data may make it impossible to include images of any aspect ratios, and the grounding ability may be difficult to generalize well to unseen aspect ratios. After being fine-tuned on the GUI-specific datasets, SeeClick improves the OCG task a lot compared to the Qwen-VL-Chat and surprisingly, it also excels at the more advanced MiniCPMV2.5. Notably, our approach, without additional SFT, substantially enhances MiniCPMV2.5's grounding ability. It achieves 84.5\% average accuracy across 10 different aspect ratio settings, outperforming MiniCPMV2.5 by 36.4\% and SeeClick by 24.3\%. 

\subsection{Task2: GUI Element Grounding}
Next, we evaluate our method on the ScreenSpot dataset which is a GUI element grounding benchmark.

\paragraph{ScreenSpot Dataset} 

It is a realistic grounding evaluation dataset proposed by~\cite{cheng2024seeclick}, which contains over 600 GUI screenshots across three platforms, i.e., mobile, desktop and web. Each screenshot contains multiple command instructions and corresponding actionable elements, which include both text and icon/widget type elements.

\paragraph{Baseline Methods}

Following~\cite{cheng2024seeclick}, we compare our method to multiple popular foundation MLLMs: MiniGPT-v2~\cite{chen2023minigpt}, Qwen-VL-Chat~\cite{qwenvl}, the latest GPT-4V~\cite{gpt4v} and MiniCPMV2.5~\cite{yao2024minicpmv}. Meanwhile, we also compare to CogAgent~\cite{hong2023cogagent} and SeeClick~\cite{cheng2024seeclick} which are SOTA GUI element grounding models supervised fine-tuned on a large amount of GUI-specific grounding tasks. To have a fair comparison, we directly use the evaluation setup in SeeClick and compare to the numbers reported in SeeClick paper. The prompt templates used for MiniCPMV2.5 and our method are presented in Fig.~\ref{fig:screenspot}. 

\paragraph{Results} 
As Table~\ref{tab:screenspot} illustrates, foundation MLLMs generally perform poorly on GUI element grounding. MiniGPT-v2 and Qwen-VL-Chat average below 6\% accuracy across platforms, while GPT-4V reaches only 16.2\%. MiniCPMV2.5 performs better at 36.0\%, likely due to OCR-related pretraining. GUI-specific fine-tuned models like CogAgent and SeeClick outperform these. Our approach, built on MiniCPMV2.5 without additional fine-tuning, achieves the highest average accuracy of 54.8\%, surpassing even GUI-specific SFT models. It excels in text grounding, with accuracies of 88.3\%, 82.5\%, and 70.9\% for mobile, desktop, and web platforms respectively. The cases demonstrated in Fig.~\ref{fig:screenspot} suggest that adaptively selecting text tokens from generated element descriptions can be more effective for GUI grounding than using query text directly.

\input{tables/tab_ablation_method_component}

\subsection{Task3: GUI Agent Evaluation}
We further evaluate our method on GUI agent benchmark.
\paragraph{Mind2Web Dataset}

\cite{deng2024mind2web} introduced the Mind2Web dataset to evaluate GUI agents in web environments using text-based HTML content. Each sample in the dataset typically consists of an open-ended, high-level goal instruction and a human action trajectory sequence, including clicking, selecting, and typing actions. While the released dataset also includes GUI screenshots corresponding to each sample, we follow~\cite{cheng2024seeclick} and evaluate our method using only the GUI images. Since this work mainly focuses on the GUI grounding task, we evaluate compared methods on the Element accuracy metric. A prediction is correct if the predicted coordinate is within the target element's bounding box for vision-based methods.

\paragraph{Baseline Methods}

We compare our method to two vision-based GUI agents Qwen-VL and SeeClick~\cite{cheng2024seeclick}, which are both fine-tuned on the Mind2Web training set. Additionally, we include the foundation MLLM MiniCPMV2.5 for comparison. The prompt template\footnote{The prompt template demonstrates our attention-driven grounding method for GUI agents, with the potential for further performance improvements through refined prompting.} used for our method is presented in Fig.~\ref{fig:mind2web_agent}. Due to space constraints, the prompt for MiniCPMV2.5, which is similar to ours, is provided in the supplementary materials.

\paragraph{Results}

Results in Table~\ref{tab:mind2web} demonstrate that the proposed attention-driven grounding method improves MiniCPMV2.5's element accuracy across all settings, achieving comparable average accuracy to the best tuning-based approach.
Fig.~\ref{fig:mind2web_agent} showcases one example that our method grounds precisely at each step and successfully achieves the overall goal.

\subsection{Ablation Study}

We investigate the impact of each component in our TAG method. In Table~\ref{tab:ablation_method_component}, adding attention-driven grounding significantly improves performance, with Mobile Text accuracy increasing from 40.3\% to 71.8\%. Introducing adaptive text token selection further enhances results, particularly for Mobile Text (86.4\%) and Icon/Widget (28.4\%). The full model, incorporating self-attention selection, achieves the best performance across all metrics, with notable improvements in Mobile Text (88.3\%) and Desktop Icon/Widget (28.6\%).

\input{figures/fig_ablation_topk_lb}

\subsection{More Discussions}

\paragraph{Impact of Top $K$}
$K$ is used for filtering self-attention weights and keeping top-ranked attentions for text-to-image mapping. Figure~\ref{fig:abl_topk} shows that reducing $K$ initially improves performance, with optimal results at $K=10$ for both aspect ratios. However, extreme values ($K=1$ or $K=1024$, i.e., not reduced) lead to decreased accuracy. This demonstrates the benefit of filtering noisy attention heads while retaining sufficient information for text-to-image mapping. Based on these results, we use $K=10$ in all experiments.

\paragraph{Impact of Threshold $\delta$}
$\delta$ is defined to determine the highlight region for final grounding prediction. Figure~\ref{fig:abl_lb} shows that with a lower threshold $\delta\leq 0.3$, the model's performance is suboptimal due to including too many fairly attended regions. As $\delta$ increases, the model's performance reaches its peak at $\delta=0.5$, but diminishes if $\delta$ is increased further. Thus $\delta=0.5$ is used across all datasets.

\paragraph{Generalization Ability}
We applied our attention-driven grounding to another foundation MLLM, Qwen-VL-Chat~\cite{qwenvl}, to demonstrate its generalization. Despite Qwen-VL-Chat's initial poor performance in GUI grounding, our method improved its accuracy from 2.7\% to 10.2\% on the 4:3 aspect ratio on our Mind2Web-OCG dataset. This showcases the broad applicability of our proposed mechanism across different foundation MLLMs.

%% file: tables/tab_m2w_loc_v1.tex
\begin{table*}[t!]
\centering
\tabcolsep=0.05cm
{\fontsize{10pt}{12pt}\selectfont
\resizebox{\textwidth}{!}{%
\begin{tabular}{lcccccccccccc}
\hline
\multirow{2}{*}{MLLMs} & \multirow{2}{*}{\parbox{1.5cm}{\centering w/o \\ SFT}} & \multicolumn{10}{c}{Aspect Ratio of Input Image (width:height)} & \multirow{2}{*}{Average} \\ \cline{3-12}
 & & 1:4 & 9:21 & 9:19 & 1:2 & 9:16 & 4:3 & 16:9 & 2:1 & 21:9 & 4:1 & \\ 
\hline
Qwen-VL-Chat & \textcolor{darkgreen}{\ding{51}} & 7.3\% & 3.2\% & 3.1\% & 2.8\% & 2.2\% & 2.7\% & 2.9\% & 3.8\% & 4.5\% & 9.7\% & 4.2\% \\ 
MiniCPMV2.5 & \textcolor{darkgreen}{\ding{51}} & 17.2\% & 13.6\% & 15.9\% & 21.4\% & 31.0\% & 80.2\% & 84.6\% & 81.1\% & 77.2\% & 59.2\% & 48.1\%\\
\textbf{TAG (Ours)} & \textcolor{darkgreen}{\ding{51}}& \textbf{86.1\%} & \textbf{80.3\%} & \textbf{80.2\%} & \textbf{84.8\%} & \textbf{84.7\%} & \textbf{82.6\%} & \textbf{86.6\%} & \textbf{87.9\%} & \textbf{83.9\%} & \textbf{88.0\%} & \textbf{84.5\%}\\
\hline
\rowcolor[gray]{0.9}
SeeClick & \textcolor{darkred}{\ding{55}} & 52.7\% & 57.5\% & 56.6\% & 56.3\% & 57.5\% & 56.6\% & 63.6\% & 66.1\% & 65.9\% & 69.1\% & 60.2\% \\
\hline
\end{tabular}%
}
}
\caption{Method comparison on the proposed OCG dataset. Our method significantly outperforms other tuning-free and tuning-based methods across all aspect ratios.
}
\label{tab:mind2web_ocg}
\vspace{-0.7em}
\end{table*}

%% file: tables/tab_screenspot.tex
\begin{table*}[t!]
\centering
\tabcolsep=0.05cm
{\fontsize{10pt}{12pt}\selectfont
\begin{tabular}{lccccccccc}
\hline
\multirow{2}{*}{MLLMs} & \multirow{2}{*}{\parbox{1.5cm}{\centering Model\\Size}} & \multirow{2}{*}{\parbox{1.5cm}{\centering w/o \\ SFT}} & \multicolumn{2}{c}{Mobile} & \multicolumn{2}{c}{Desktop} & \multicolumn{2}{c}{Web} & \multirow{2}{*}{Average} \\ \cline{4-9}
 & & & Text & Icon/Widget & Text & Icon/Widget & Text & Icon/Widget & \\ 
\hline
MiniGPT-v2 & 7B & \textcolor{darkgreen}{\ding{51}} & 8.4\% & 6.6\% & 6.2\% & 2.9\% & 6.5\% & 3.4\% & 5.7\% \\
Qwen-VL-Chat & 9.6B & \textcolor{darkgreen}{\ding{51}} & 9.5\% & 4.8\% & 5.7\% & 5.0\% & 3.5\% & 2.4\% & 5.2\% \\ 
GPT-4V & - & \textcolor{darkgreen}{\ding{51}} & 22.6\% & 24.5\% & 20.2\% & 11.8\% &9.2\% & 8.8\% & 16.2\% \\
MiniCPMV2.5 & 8.5B & \textcolor{darkgreen}{\ding{51}} & 40.3\% & 14.0\% & 62.4\% & 12.1\% & 67.4\% & 19.9\% & 36.0\% \\
\textbf{TAG (Ours)} & 8.5B & \textcolor{darkgreen}{\ding{51}}& \textbf{88.3\%} & \textbf{29.3\%} & \textbf{82.5}\% & \textbf{28.6\%} & \textbf{70.9\%} & \textbf{29.1\%} & \textbf{54.8\%} \\
\hline
\rowcolor[gray]{0.9} CogAgent & 18B & \textcolor{darkred}{\ding{55}} & 67.0\% & 24.0\% & \textbf{74.2\%} & 20.0\% & \textbf{70.4\%} & 28.6\% & 47.4\% \\
\rowcolor[gray]{0.9}
SeeClick & 9.6B & \textcolor{darkred}{\ding{55}} & \textbf{78.0\%} & \textbf{52.0\%} & 72.2\% & \textbf{30.0\%} & 55.7\% & \textbf{32.5\%} & \textbf{53.4\%} \\ 
\hline
\end{tabular}%
}
\caption{Method comparison on Screenspot. The highest value in each column is bolded, considering both the upper section of tuning-free approaches and the lower section of tuning-based approaches.
}
\label{tab:screenspot}
\vspace{-0.7em}
\end{table*}

%% file: figures/fig_screenspot.tex
\begin{figure*}[t!]
\centering
\includegraphics[width=0.85\textwidth]{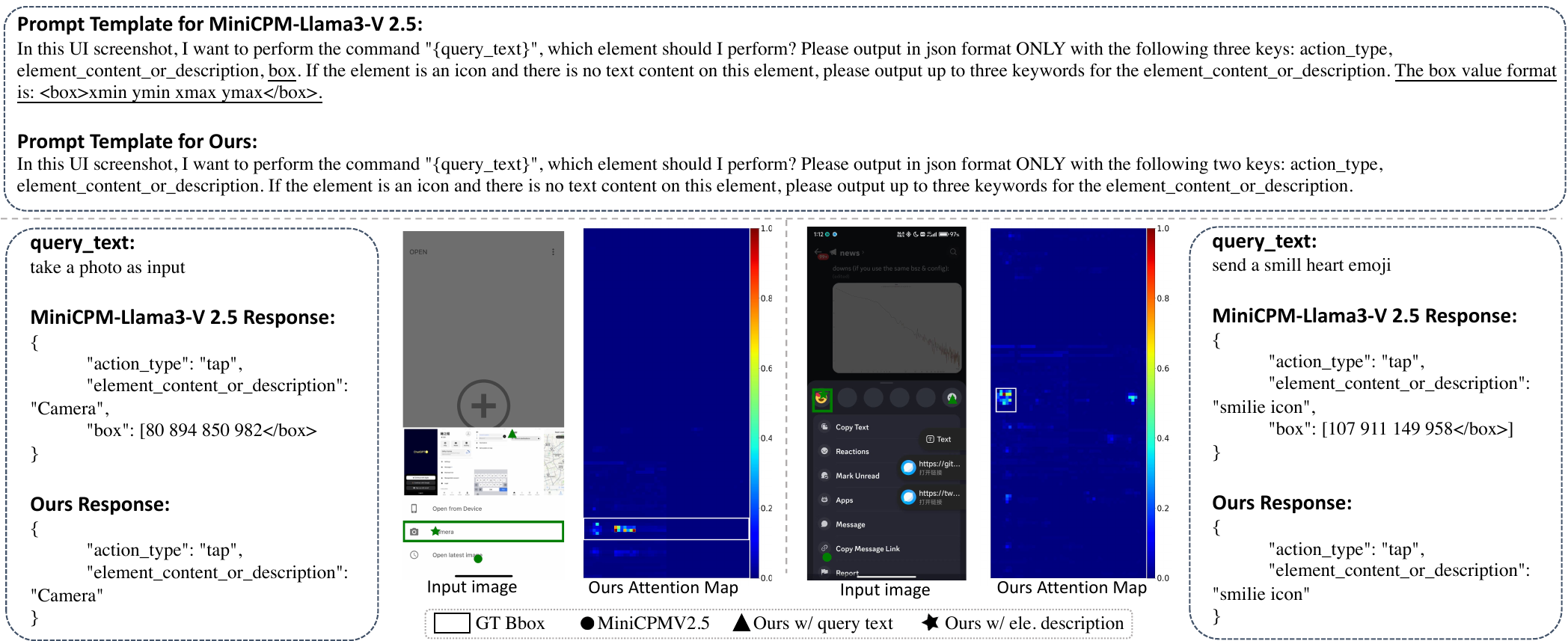}
\captionsetup{skip=2pt}
\caption{Demonstration of the comparing methods on two cases of ScreenSpot. Our attention-driven grounding with element description success in localizing the text and icon elements respectively. Please zoom in for a better view.}
\label{fig:screenspot}
\vspace{-0.65cm}
\end{figure*}

%% file: figures/fig_mind2web_agent.tex
\begin{figure*}[t!]
\centering
\includegraphics[width=0.85\textwidth]{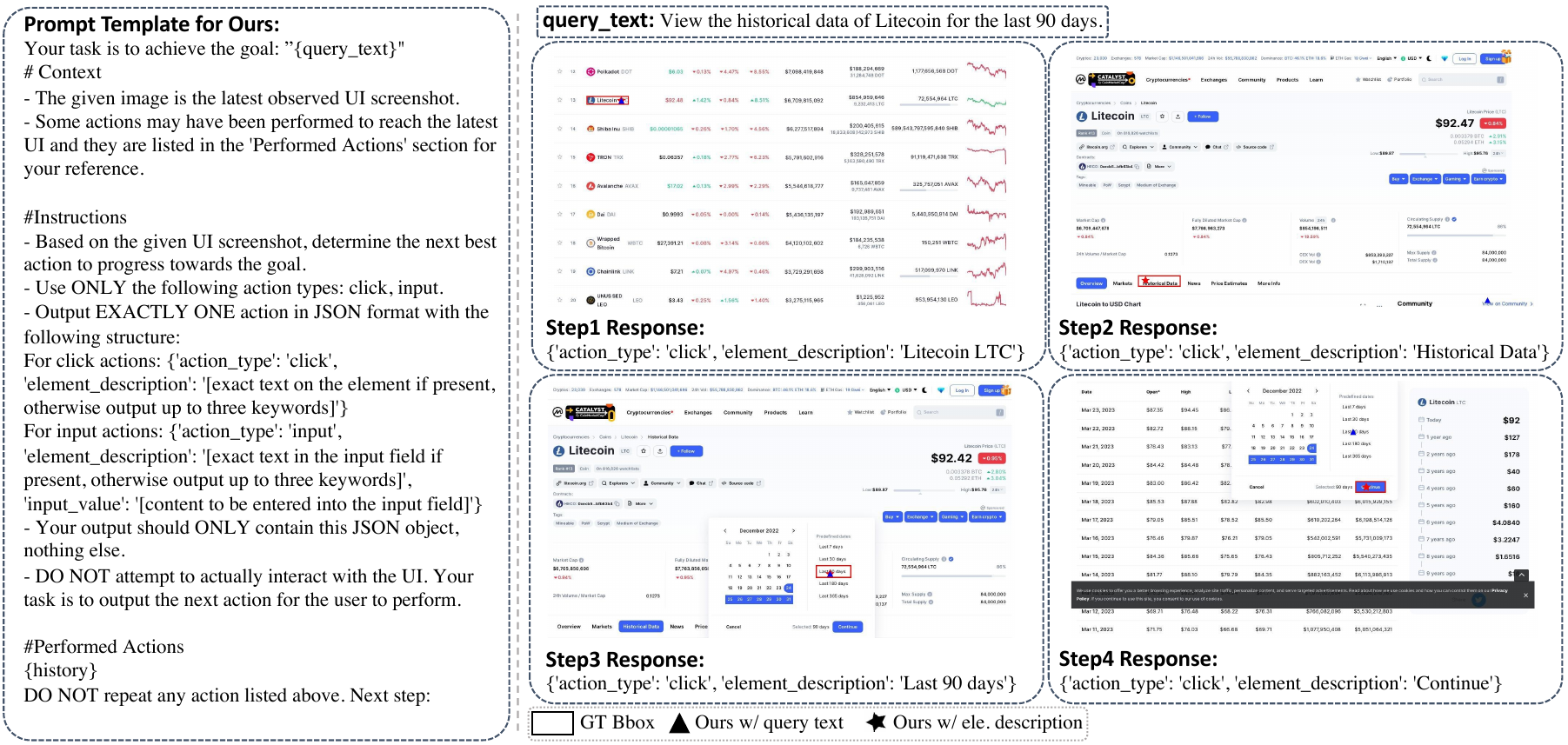}
\captionsetup{skip=2pt}
\caption{Demonstration of our method on Mind2Web to ground precisely at each step and successfully achieve the overall goal. Detailed action history is presented in supplementary materials. Please zoom in for a better view.
}
\label{fig:mind2web_agent}
\vspace{-1.0em}
\end{figure*}

%% file: tables/tab_m2w_agent_v2.tex
\begin{table}[t!]
\centering
\renewcommand\arraystretch{1.2}
\tabcolsep=0.10cm
{\fontsize{10pt}{12pt}\selectfont
\begin{tabular}{lccccc}
\hline
\multirow{2}{*}{\makecell{MLLMs}} & \multirow{2}{*}{\makecell{w/o \\ SFT}} & \multirow{2}{*}{\makecell{Cross-\\Task}} & \multirow{2}{*}{\makecell{Cross-\\Website}} & \multirow{2}{*}{\makecell{Cross-\\Domain}} & \multirow{2}{*}{\makecell{Average}} \\ 
& & & & &\\
\hline
MiniCPMV2.5 & \textcolor{darkgreen}{\ding{51}} & 15.0\% & 13.8\% & 18.2\% & 15.7\%\\
\textbf{TAG (Ours)} & \textcolor{darkgreen}{\ding{51}} & 25.4\% & 20.6\% & \textbf{26.8\%} & \textbf{24.3\%} \\
\hline
\rowcolor[gray]{0.9}
Qwen-VL$^*$ & \textcolor{darkred}{\ding{55}} & 15.9\% & 13.2\% & 14.1\% & 14.4\% \\
\rowcolor[gray]{0.9}
SeeClick$^*$ & \textcolor{darkred}{\ding{55}} & \textbf{28.3\%} & \textbf{21.4\%} & 23.2\% & \textbf{24.3\%} \\
\hline
\end{tabular}%
}
\caption{Element accuracy on Mind2Web dataset. The highest value in each column is bolded. Qwen-VL$^*$ and SeeClick$^*$ refer to the fine-tuning of Qwen-VL-Chat and SeeClick models, respectively, on Mind2Web training set.
}
\vspace{-1.75em}
\label{tab:mind2web}
\end{table}

%% file: tables/tab_ablation_method_component.tex
\begin{table}[t!]
\centering
\tabcolsep=0.05cm
{\fontsize{10pt}{12pt}\selectfont
\resizebox{\columnwidth}{!}{%
\begin{tabular}{ccccccc}
\hline
\multirow{2}{*}{\parbox{1.5cm}{\centering Attn-d.\\ground}} & \multirow{2}{*}{\parbox{1.5cm}{\centering Token \\Select}} & \multirow{2}{*}{\parbox{1.5cm}{\centering self-attn\\filtering}} & \multicolumn{2}{c}{Mobile} & \multicolumn{2}{c}{Desktop} \\ 
\cline{4-7}
 & & & Text & Icon/W. & Text & Icon/W. \\ 
\hline
\textcolor{darkred}{\ding{55}} & \textcolor{darkred}{\ding{55}} & \textcolor{darkred}{\ding{55}} & 40.3\% & 14.0\% & 62.4\% & 12.1\% \\
\textcolor{darkgreen}{\ding{51}} & \textcolor{darkred}{\ding{55}} & \textcolor{darkred}{\ding{55}} & 71.8\% & 27.1\% & 73.2\% & 20.0\% \\ 
\textcolor{darkgreen}{\ding{51}} & \textcolor{darkgreen}{\ding{51}} & \textcolor{darkred}{\ding{55}} & 86.4\% & 28.4\% & 77.3\% & 24.3\% \\
\textcolor{darkgreen}{\ding{51}} & \textcolor{darkred}{\ding{55}} & \textcolor{darkgreen}{\ding{51}} & 80.9\% & 27.5\% & 78.8\% & 25.0\% \\
\textcolor{darkgreen}{\ding{51}} & \textcolor{darkgreen}{\ding{51}} & \textcolor{darkgreen}{\ding{51}} & \textbf{88.3\%} & \textbf{29.3\%} & \textbf{82.5}\% & \textbf{28.6\%} \\
\hline
\end{tabular}%
}
}
\caption{Ablation on each component of our method.
}
\label{tab:ablation_method_component}
\vspace{-1.75em}
\end{table}

%% file: figures/fig_ablation_topk_lb.tex
\begin{figure}[t]
  \centering
  \begin{minipage}[b]{0.49\columnwidth}
    \centering
    \includegraphics[width=\textwidth]{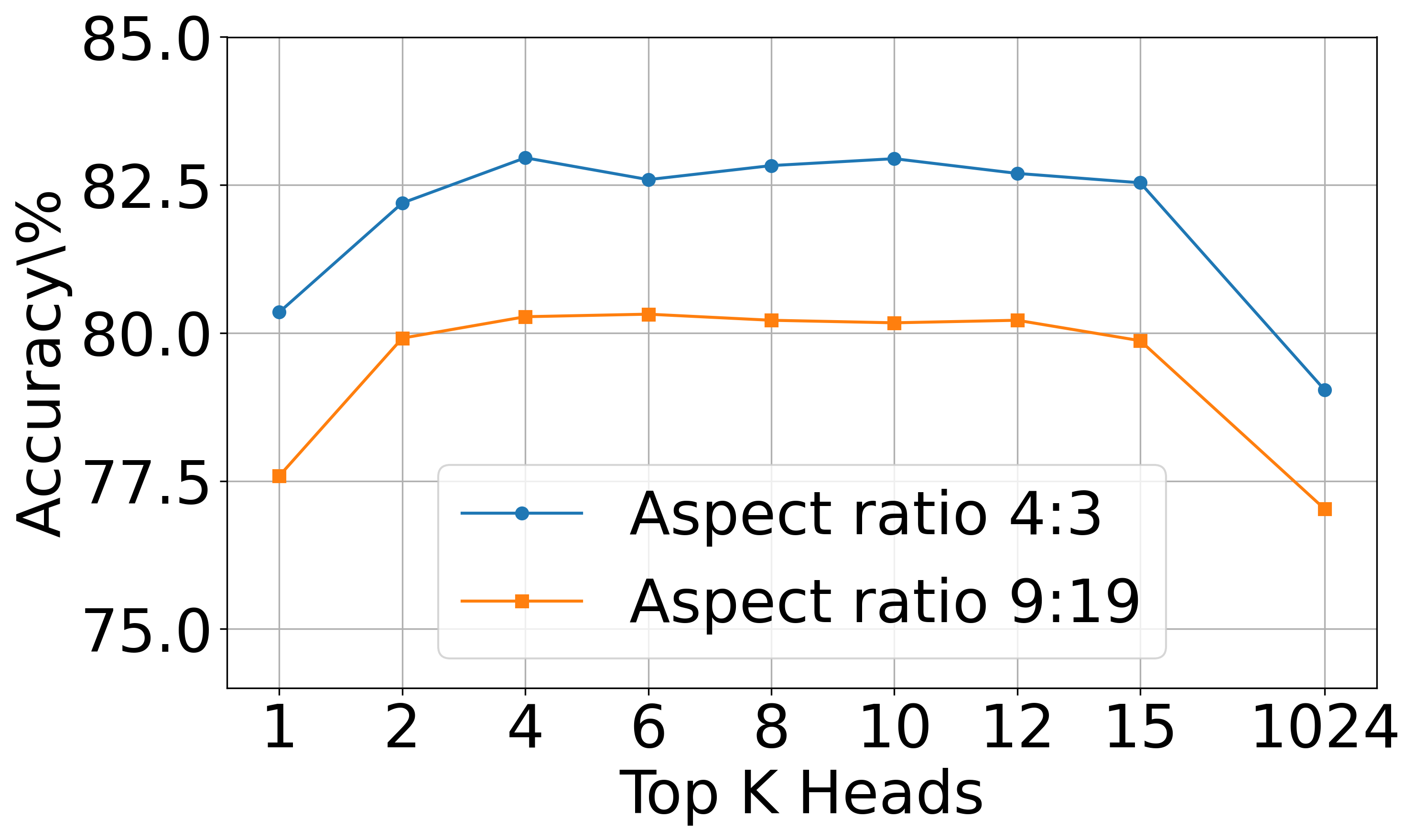}
    \caption{Ablation on Top $K$.}
    \label{fig:abl_topk}
  \end{minipage}
  \hfill
  \begin{minipage}[b]{0.49\columnwidth}
    \centering
    \includegraphics[width=\textwidth]{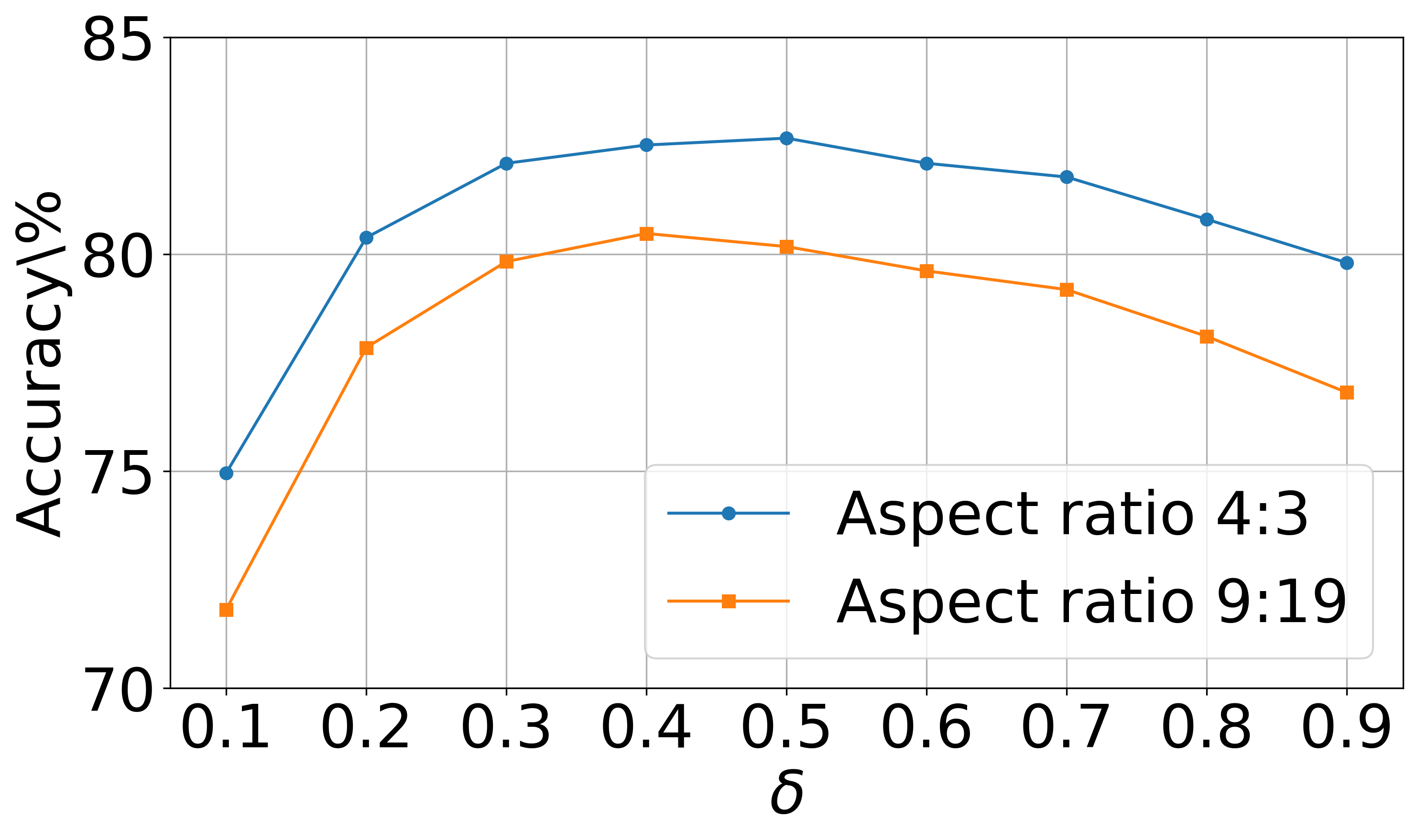}
    \caption{Ablation on $\delta$.}
    \label{fig:abl_lb}
  \end{minipage}
  \vspace{-1.75em}
\end{figure}

%% file: contents/5-conclusion.tex
\section{Conclusion}
In this paper, we introduce the Tuning-free Attention-driven Grounding (TAG) method, which uses the inherent attention mechanisms of pretrained MLLMs to accurately ground GUI elements without additional fine-tuning. Applied to the MiniCPM-Llama3-V 2.5 model, TAG demonstrates that leveraging built-in model capabilities can effectively match or exceed the performance of traditional methods, particularly in text localization tasks. These suggest that MLLMs can be used more efficiently, reducing the need for resource-intensive fine-tuning while avoiding the risk of overfitting. TAG has the potential to be applied across various models and multimodal scenarios, offering a promising method for enhancing AI's adaptability in interacting with user interfaces.

\paragraph{Limitations}
TAG relies heavily on the capabilities and the quality of the pretrained models it uses. If these models have inherent biases or have not been trained on diverse enough data, TAG's effectiveness could be limited, potentially affecting its accuracy and generalization ability. To alleviate this, we can expand the training datasets used for pretraining the MLLMs, which, while promising, is beyond the scope of this paper. We regard it as our future work.

%% file: contents/7-supp.tex
\section{Supplementary Materials}
This supplementary material provides additional details to complement our main manuscript. It includes the following sections:
\begin{enumerate}
    \item \textbf{Full Output of Query in Figure 1 of Main Paper (Section \ref{sec:full_output})} Full output of query prompt presented in Figure 1 of main paper.
    \item \textbf{MiniCPMV2.5 for Grounding (Section \ref{sec:minicpmv2.5_for_grounding})}: Detailed explanation of how MiniCPMV2.5 is utilized for grounding tasks.
    \item \textbf{OCG Dataset (Section \ref{sec:mind2web-ocg})}: Statistical information and characteristics of our proposed \textit{OCG} dataset.
    \item \textbf{Prompt Template for MiniCPMV2.5 (Section \ref{sec:prompt_temp_minicpmv_m2w})}: The specific prompt template used with MiniCPMV2.5 on the Mind2Web agent dataset.
    \item \textbf{Action History for Mind2Web Dataset (Section \ref{sec:history})}: Description of the action history content used for the Mind2Web agent dataset.
    \item \textbf{Inference Efficiency Analysis for TAG (Section \ref{sec:tag_inference_efficiency})}: Comparison of inference time costs between our proposed TAG and the MiniCPMV2.5 model, highlighting efficiency improvements.
    \item \textbf{Evaluation on VisualWebBench (Section \ref{sec:visualwebbench_evaluation})}: Performance comparison of TAG and baseline methods on the Element Grounding task in VisualWebBench.
\end{enumerate}

\subsection{Full Output of Query in Figure 1 of Main Paper}\label{sec:full_output}
Due to space limitations, Figure 1 in the main paper only shows partial results of the model outputs. Figure~\ref{fig:full_output} in this supplementary material presents the complete outputs. Specifically, we conducted an experiment to compare the optical character grounding abilities of MiniCPMV2.5 and our TAG method. First, we used MiniCPMV2.5 to perform OCR on an image, with the prompt ``\texttt{Please do OCR on this image.}'' We then used the generated text as query inputs for both models to test their text grounding capabilities.

For MiniCPMV2.5, we used the prompt: ``\texttt{What is the bounding box of "{query\_text}" in the image? The bounding box output format is: \textless box\textgreater xmin ymin xmax ymax\textless/box\textgreater. Please directly output the bounding box.}'' For our TAG method, we used a similar prompt: ``\texttt{What is the bounding box of "{query\_text}"}''.

We presented six examples, each marked in a distinct color. The results demonstrated that while MiniCPMV2.5 could understand the text, it failed to accurately ground the query text in the image. In contrast, our TAG method successfully grounded the text precisely, showing superior performance in this task.

\subsection{MiniCPMV2.5 for Grounding}\label{sec:minicpmv2.5_for_grounding}
MiniCPMV2.5 has been pre-trained on OCR-related datasets, incorporating three location formats according to the author's feedback\footnote{\url{https://github.com/OpenBMB/MiniCPM-V/issues/185#issuecomment-2140732729}.}:
\begin{align}
    & \text{\textless point\textgreater x1 y1\textless/point\textgreater} \\
    & \text{\textless box\textgreater xmin ymin xmax ymax\textless/box\textgreater} \\
    & \text{\textless quad\textgreater x1 y1 x2 y2 x3 y3 x4 y4\textless/quad\textgreater}
\end{align}
Our empirical observations indicate that MiniCPMV2.5 tends to output position responses in the box format. To facilitate consistent parsing, we include the box format in the prompt, guiding MiniCPMV2.5 to generate stable, easily interpretable locations.

During MiniCPMV2.5's pre-training, location coordinates were rescaled to the range [0, 1000]\footnote{\url{https://github.com/OpenBMB/MiniCPM-V/issues/185#issuecomment-2141217036}.}. To map these generated coordinates back to the input image space, we apply the following transformations:
\begin{align}
    \text{xmin}' & = \text{xmin} \cdot W/1000 \\
    \text{ymin}' & = \text{ymin} \cdot H/1000 \\
    \text{xmax}' & = \text{xmax} \cdot W/1000 \\
    \text{ymax}' & = \text{ymax} \cdot H/1000
\end{align}
Where $W$ and $H$ represent the width and height of the input image, respectively.
For the final grounding prediction, we calculate the center of the transformed bounding box. As demonstrated in Table 1 of our main paper, this approach enables MiniCPMV2.5 to achieve a descent grounding accuracy.

\subsection{OCG Dataset}\label{sec:mind2web-ocg}
\textit{OCG} is an optical character grounding dataset derived from the Mind2Web test set. It is designed to validate the text-to-image mapping capability of the proposed attention-driven grounding approach. The dataset comprises 104 website page screenshots, each accompanied by OCR data (including text and corresponding bounding boxes) obtained using the Azure Vision API.
To accommodate various aspect ratios, images are cropped from the original screenshots while preserving all text and bounding boxes that fall entirely within the cropped area. Consequently, the number of text elements to be grounded varies across different aspect ratio settings. Detailed statistical information about the dataset is presented in Table~\ref{tab:ocg_stats}.

\input{tables/tab_ocg_stats}

\subsection{Prompt Template for MiniCPMV2.5 on Mind2Web Dataset}\label{sec:prompt_temp_minicpmv_m2w}
The prompt template used for MiniCPMV2.5 is presented in Figure~\ref{fig:prompt_minicpmv2.5_m2w}. It is almost the same as the templates used for our TAG method for a fair comparison. 

\input{figures/fig_supp_full_output}
\input{figures/fig_supp_prompt_minicpmv2.5_m2w}
\input{figures/fig_supp_action_history}

\subsection{Action History for Mind2Web Dataset}\label{sec:history}
The Mind2Web dataset comprises samples with overall goals that require multiple-step interactions with the GUI to accomplish. To provide context on how the current state was reached, historical actions is typically included. Following the approach used in SeeClick~\cite{cheng2024seeclick}, we incorporate the latest four ground-truth steps in the prompt to ensure a fair comparison. Taking the case shown in Figure 5 of our main paper as an example, the action history used for each step is presented in Figure~\ref{fig:action_history}.

\subsection{Inference Efficiency Analysis for TAG}\label{sec:tag_inference_efficiency}
Table~\ref{tab:inference_time} compares the average computation time for TAG and MiniCPMV2.5 on the screenspot-desktop test split. The prompt used is as shown in Figure 4 of the main paper. While TAG adds slightly more computations with attention selection and product operations, it avoids outputting bounding boxes, reducing token length during inference and potentially lowering overall time cost. Results are obtained on NVIDIA RTX 4090 GPU.

\input{tables/tab_supp_inference_time}

\subsection{Evaluation on VisualWebBench}\label{sec:visualwebbench_evaluation}
We further evaluate our method on the Element Grounding task of another benchmark VisualWebBench~\cite{liu2024visualwebbench}. Its images are marked with bounding boxes and tags on seven potential elements, allowing MLLM to perform multi-choice (MC) prediction. For MiniCPMV2.5, we evaluate multi-choice grounding and direct bounding box prediction. In Table~\ref{tab:visualwebbench}, TAG outperforms all models.

\input{tables/tab_supp_visualwebbench}

%% file: tables/tab_ocg_stats.tex
\begin{table}[H]
\centering
\tabcolsep=0.05cm
{\fontsize{10pt}{12pt}\selectfont
\begin{tabular}{|c|c|c|c|c|c|c|c|c|c|}
\hline
1:4 & 9:21 & 9:19 & 1:2 & 9:16 & 4:3 & 16:9 & 2:1 & 21:9 & 4:1 \\ 
\hline
1030 & 2066 & 2320 & 2441 & 2811 & 3797 & 2904 & 2602 & 2267 & 1489 \\ 
\hline
\end{tabular}%
}
\caption{Number of OCG samples of each aspect ratio setting (width:height).
}
\label{tab:ocg_stats}
\end{table}

%% file: figures/fig_supp_full_output.tex
\begin{figure*}[t!]
\centering
\includegraphics[width=\textwidth]{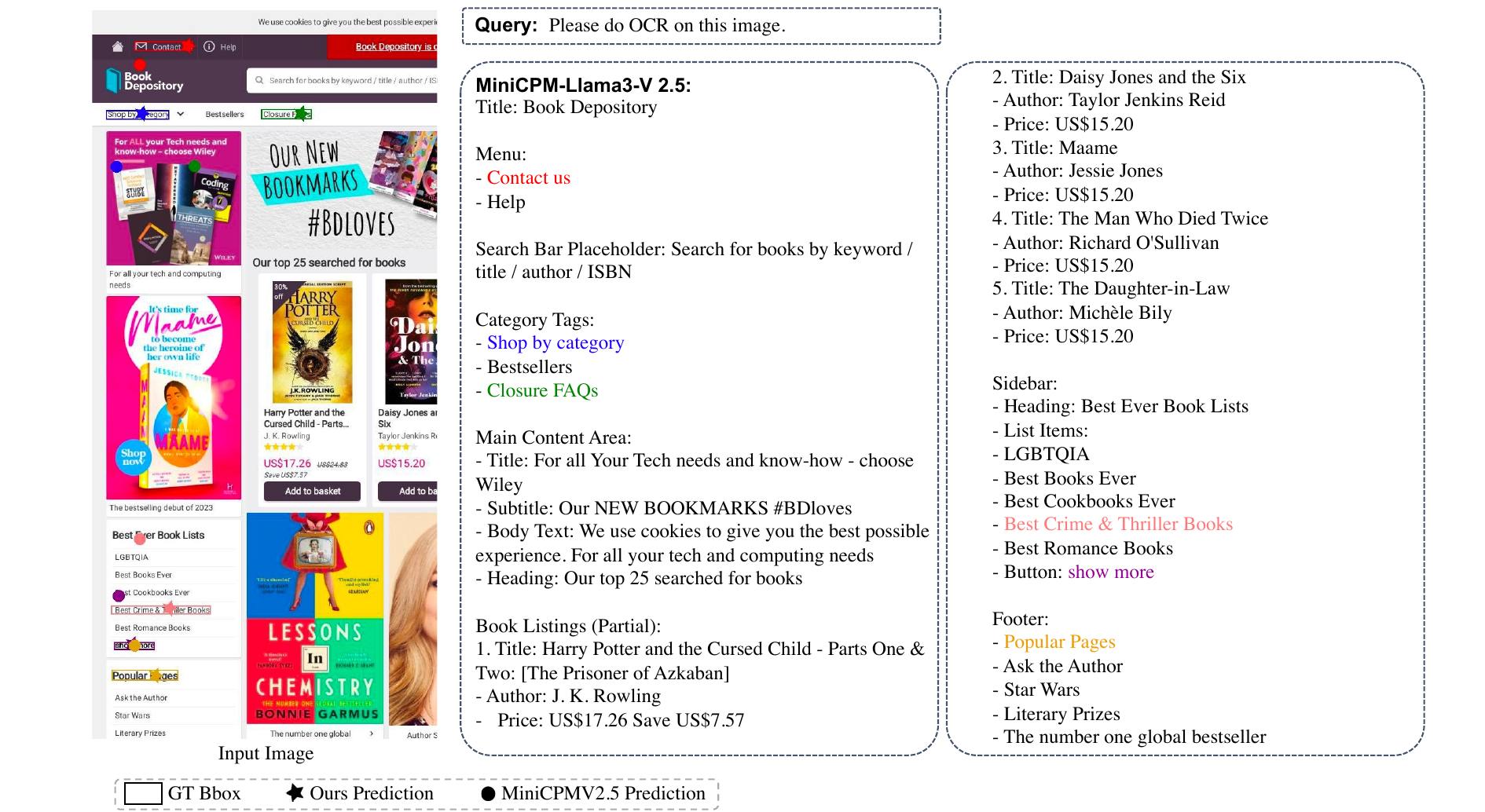}
\captionsetup{skip=2pt}
\caption{
Demonstration of the Full output for the query ``Please do OCR on this image.'' with MiniCPMV2.5.
}
\label{fig:full_output}
\end{figure*}

%% file: figures/fig_supp_prompt_minicpmv2.5_m2w.tex
\begin{figure*}[t!]
\centering
\includegraphics[width=\textwidth]{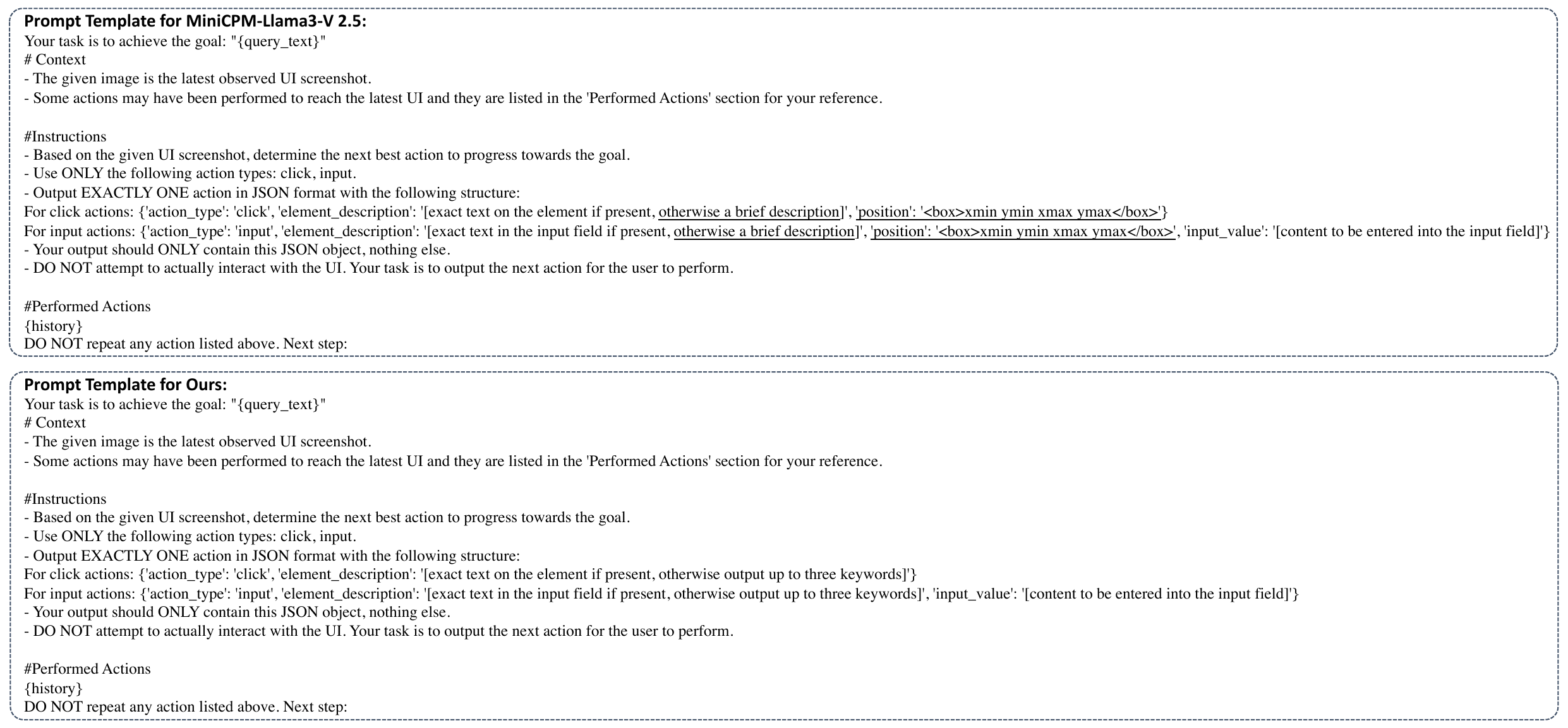}
\captionsetup{skip=2pt}
\caption{
The prompt templates used for MiniCPM-Llama3-V 2.5 and our model to accomplish tasks in the Mind2Web agent dataset. These two prompt templates are largely similar and the main distinctions are underlined.
}
\label{fig:prompt_minicpmv2.5_m2w}
\end{figure*}

%% file: figures/fig_supp_action_history.tex
\begin{figure*}[t!]
\centering
\includegraphics[width=\textwidth]{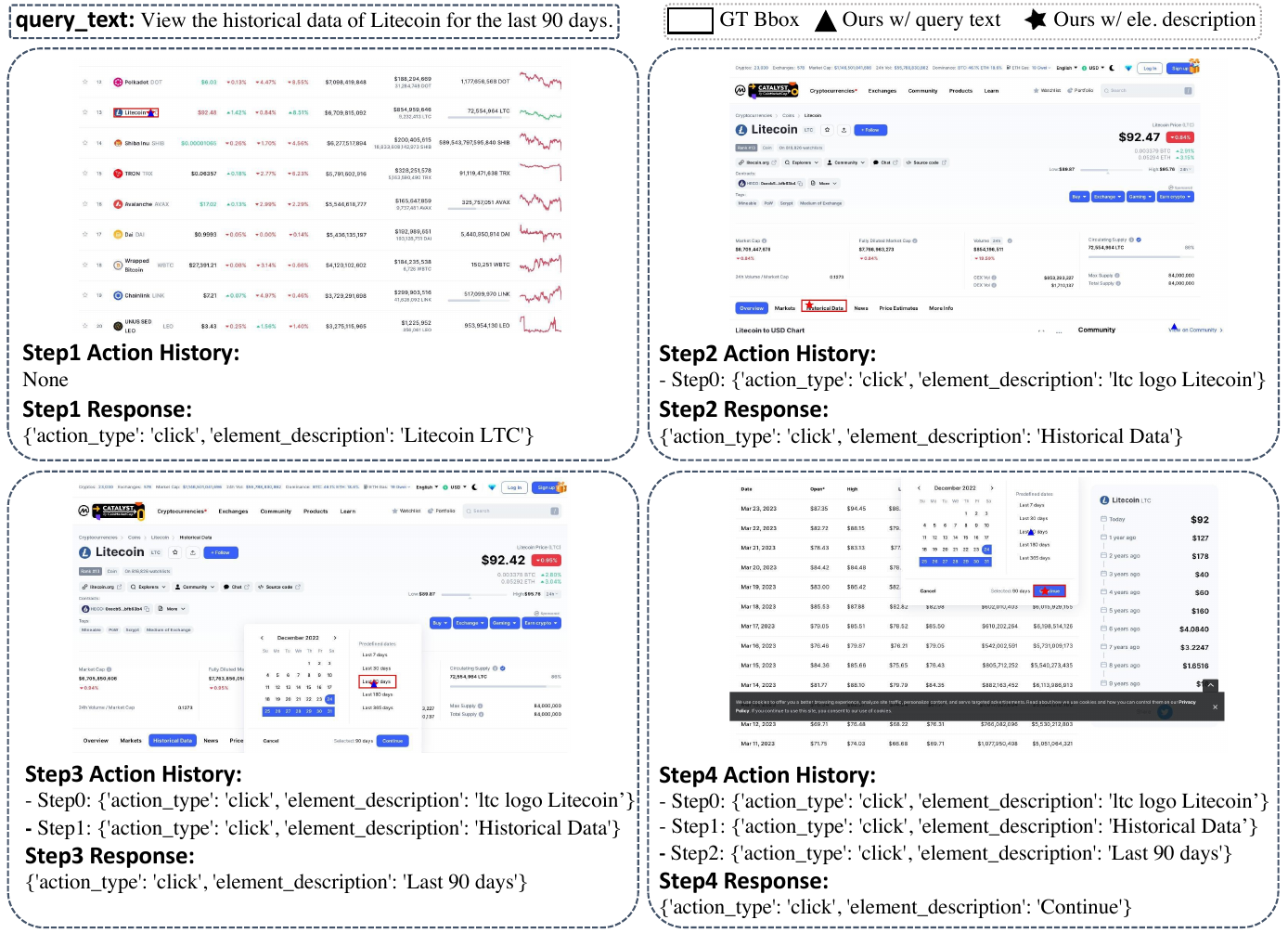}
\captionsetup{skip=2pt}
\caption{
Demonstration of our method on Mind2Web to ground precisely at each step and successfully achieve the overall goal with the latest four ground-truth history actions.
}
\label{fig:action_history}
\end{figure*}

%% file: tables/tab_supp_inference_time.tex
\begin{table}[H]
\centering
\tabcolsep=0.05cm
{\fontsize{10pt}{12pt}\selectfont
\begin{tabular}{|c|c|c|c|}
\hline
\multirow{2}{*}{MLLMs} & \multicolumn{3}{c|}{ScreenSpot-Desktop} \\
\cline{2-4}
 & Text$\uparrow$ & Icon/Widget$\uparrow$ & Time Cost (s)$\downarrow$ \\ 
\hline
MiniCPMV2.5 & 62.4\% & 12.1\% & 1.10 \\
\textbf{TAG (Ours)} & \textbf{82.5\%} & \textbf{28.6\%} & \textbf{1.04} \\
\hline
\end{tabular}%
}
\caption{Average Inference time cost comparison between TAG and MiniCPMV2.5.
}
\label{tab:inference_time}
\end{table}

%% file: tables/tab_supp_visualwebbench.tex
\begin{table}[H]
\centering
\tabcolsep=0.05cm
{\fontsize{10pt}{12pt}\selectfont
\begin{tabular}{|l|c|}
\hline
MLLMs & Element Grounding \\
\hline
Claude Sonnet-MC & 68.8\%\\
SeeClick-MC & 41.6\%\\
MiniCPMV2.5-MC & 17.2\% \\
MiniCPMV2.5 & 78.4\% \\
\textbf{TAG (Ours)} & \textbf{87.4\%}\\
\hline
\end{tabular}%
}
\caption{Method comparison on the Element Grounding task in VisualWebBench benchmark. The results of methods denoted with ``-MC" are derived from the VisualWebBench paper.
}
\label{tab:visualwebbench}
\end{table}